\documentclass[preprint,review,10pt,authoryear]{elsarticle}
\usepackage{hyperref}
\usepackage{amsmath}
\usepackage{amssymb}
\usepackage{enumerate}
\usepackage{algorithm,algpseudocode}
\usepackage{verbatim} 
\usepackage{caption}
\usepackage{url}
\usepackage{footnote}
\usepackage{epstopdf}
\usepackage{epsfig}
\usepackage{changes}
\usepackage{graphicx, subfigure}
\usepackage{array}
\usepackage{amsfonts,epsfig,array,graphicx,amssymb,epstopdf}
\usepackage[]{natbib}
\usepackage{cite}

\journal{Journal of Expert Systems with Applications}

\begin{document}

\begin{frontmatter}

\title{Queuing Theory Guided Intelligent Traffic Scheduling through Video Analysis using Dirichlet Process Mixture Model}

\author{$^*$Santhosh Kelathodi Kumaran$^a$,
         Debi Prosad Dogra$^b$ and Partha Pratim Roy$^c$}
\address{School of Electrical Sciences,\\Indian Institute of Technology Bhubaneswar, Bhubaneswar-752050, India$^{a,b}$\\Department of Computer Science and Engineering,\\Indian Institute of Technology Roorkee, Roorkee-247667, India$^c$\\
Email:\ sk47@iitbbs.ac.in$^a$, dpdogra@iitbbs.ac.in$^b$,  proy.fcs@iitr.ac.in$^c$}

\begin{abstract}
Accurate prediction of traffic signal duration for roadway junction is a challenging problem due to the dynamic nature of traffic flows. Though supervised learning can be used, parameters may vary across roadway junctions. In this paper, we present a computer vision guided expert system that can learn the departure rate ($\mu$) of a given traffic junction modeled using traditional queuing theory. First, we temporally group the optical flow of the moving vehicles using Dirichlet Process Mixture Model (DPMM). These groups are referred to as tracklets or temporal clusters. Tracklet features are then used to learn the dynamic behavior of a traffic junction, especially during on/off cycles of a signal. The proposed queuing theory based approach can predict the signal open duration for the next cycle with higher accuracy when compared with other popular features used for tracking. The hypothesis has been verified on two publicly available video datasets. The results reveal that the DPMM based features are better than existing tracking frameworks to estimate $\mu$. Thus, signal duration prediction is more accurate when tested on these datasets.The method can be used for designing intelligent operator-independent traffic control systems for roadway junctions at cities and highways.
\end{abstract}

\begin{keyword}
\texttt{Traffic Intersection Management, Signal Duration Prediction, Dirichlet Process, Queuing Theory, Unsupervised Learning, Computer Vision, Visual Surveillance.}

\end{keyword}

\end{frontmatter}

\section{Introduction}
\label{sec:introduction}
Efficient traffic management is a key to handle congestions. An entire city can choke under traffic congestion if not handled carefully. Therefore, deadlock or starvation free traffic flow is the key to developing expert systems such as intelligent transportation system (ITS). As the traffic flow varies over time in a given junction, the signal management algorithms need to be adaptive. Without loss of generality, we may assume that past knowledge can help intelligent traffic management systems to adjust traffic signals accordingly. 

\begin{figure}[t]
\begin{center}
   \includegraphics[width=1.0\linewidth]{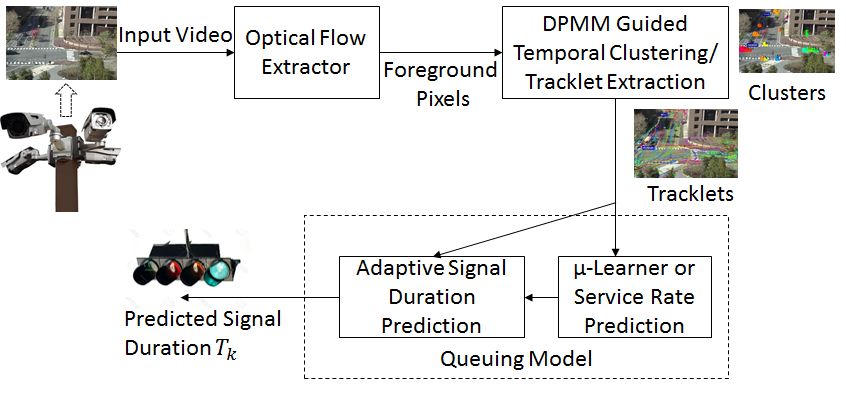}
\end{center}
   \caption{Overview of the system queuing guided traffic signal duration predictor. Initially, optical flow features extracted are fed to a temporal clustering module implemented using DPMM. In the next phase, these clusters (or tracklets) are fed to a $\mu$ learner algorithm to learn the service rate of the channel or lane. The learned $\mu$ is then used to estimate/predict the signal durations for the subsequent cycles.}
\label{fig:Overview}
\end{figure}

With the advancement of sensor technology and emergence of intelligent video surveillance systems, traffic signal management can be automated or semi-automated. In this work, we attempt to use vehicle tracklets derived using Dirichlet Process Mixture Model (DPMM)~\citep*{Rasmussen2000} based clustering method to study and understand the traffic states at junctions. Also, we model the traffic junctions using queuing theory to predict signal on/off durations for unidirectional flows. An overview of the system is presented in Fig.~\ref{fig:Overview}.

It is well-known that vehicular traffic usually follow a typical queue discipline, where vehicles move one behind the other. Though vehicles may overtake, most of the time the movement follows a first-come-first-served (FCFS) pattern. Thus, intuitively traditional queuing theory may be applied to understand the traffic state. Queuing theory has successfully been applied in many other fields such as network traffic analysis~\citep*{MLi, LCWang}, web applications~\citep*{XLiu, RTolosana-Calasanz}, scheduling~\citep*{BBensaou, AStamoulis}, etc. A queuing system is characterized by distribution of inter-arrival time, service time and the number of servers. Consider a highway traffic in steady state condition. When we watch the traffic from the top, it can be observed that the incoming  and outgoing traffic rates are same, i.e., there is no queuing in the system under normal circumstances. If we consider a junction with signal, though incoming rate may remain steady, outgoing traffic depends on the signal open/close duration. This can be logically explained as a queuing model with service rate depending on the signal duration. If this pattern is modeled, traffic states can be better interpreted and hence controlling the traffic can be done with less errors. 
    
\begin{figure}[t]
\begin{center}
 \subfigure[Queuing System]{\includegraphics[scale=0.2]{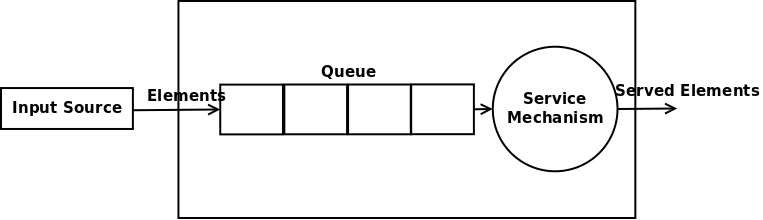}}   
 \subfigure[DPMM]{\includegraphics[scale=0.55]{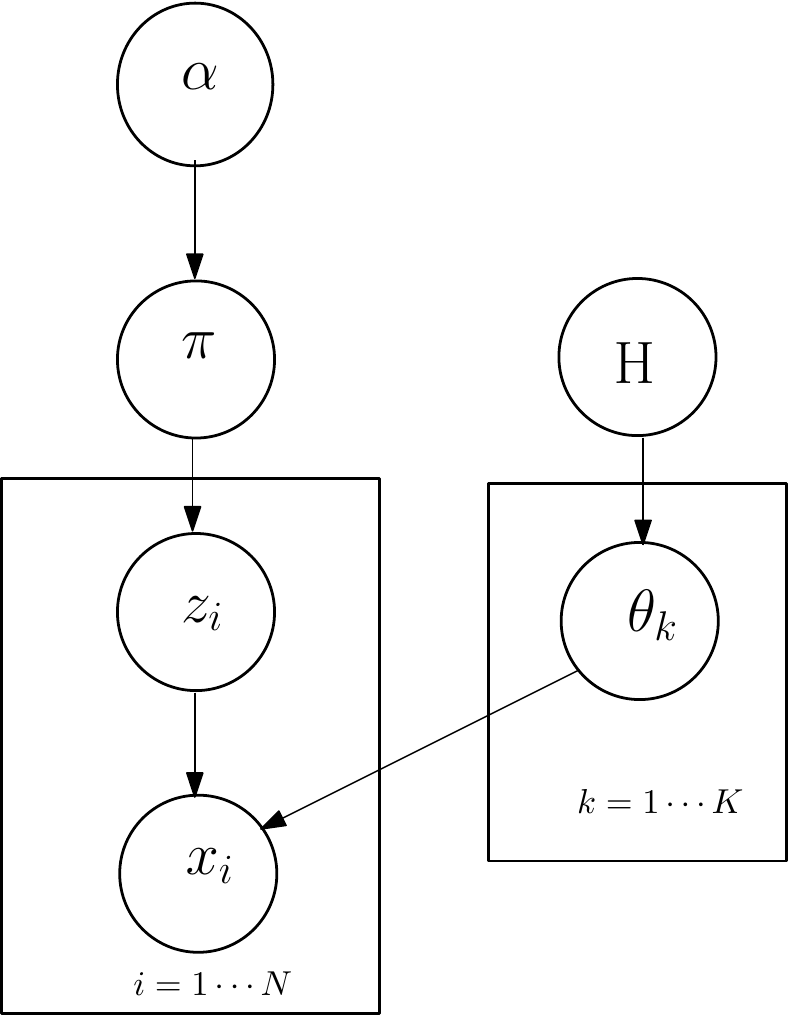}}   
\end{center}
   \caption{Base models of our proposed method. (a) A typical Queuing system. (b) A Conventional DPMM typically used in clustering of time varying data.}
\label{fig:BUILDING_BLOCK}
\end{figure}

Queuing models are typically expressed in terms of arrival rate ($\lambda$), service/departure rate ($\mu$) and the number of servers. Fig.~\ref{fig:BUILDING_BLOCK}(a) shows a typical representation of queuing system. Some of the observations on a road with traffic movement are, (i) in steady-state condition, departure rate ($\mu$) can be assumed to be same as arrival rate ($\lambda$), (ii) any change in steady state is indicated by the change in $\mu$, as traffic blockage or release is triggered at the front of the queue,  (iii) in case of traffic blockage, the trigger is from the departure point and $\mu$ gradually reduces and eventually the traffic comes to a halt and $\lambda$ needs to be controlled through alternate route planning, (iv) difference between no-traffic and traffic-blockage has to be identified from the source of the trigger, (v) in case of no traffic, trigger will be from the entry point, i.e., $\lambda$ gradually reduces and eventually settles at zero, (vi) finally, the queue length can be used to decide the signal open duration when the blockage is due to regulation of the flow. 

\begin{figure}[t]
\begin{center}
 \subfigure[Car Scene]{\includegraphics[scale=0.5]{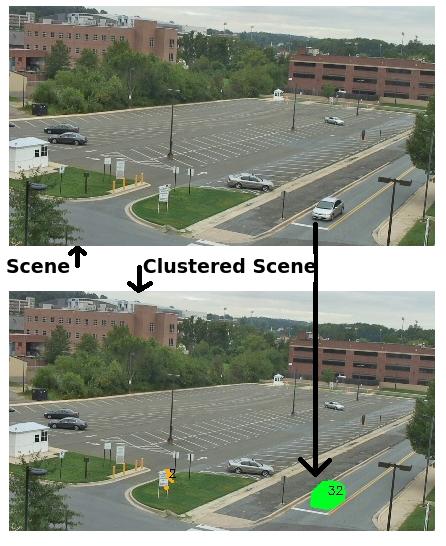}}   
 \subfigure[Bus Scene]{\includegraphics[scale=0.5]{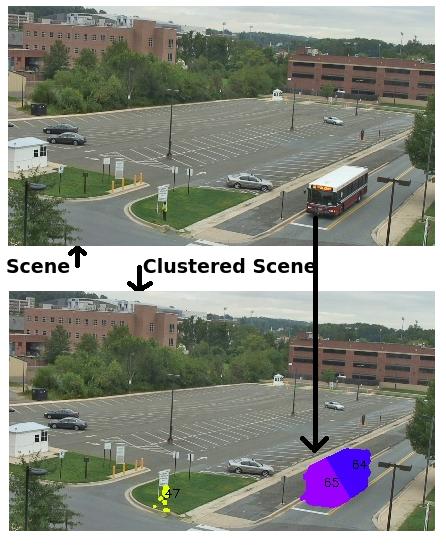}}
\end{center}
   \caption{Snapshots of clusters corresponding to a car object and a bus object. (a) A typical scene of a road segment with a car on it and cluster corresponding to the car shown in green color with an associated label. (c) A typical scene of a road segment with a bus on it and two clusters corresponding to the bus shown in blue and magenta colors with associated labels.}
\label{fig:CLUSTER_VS_VEHICLE}
\end{figure}

Based on the aforementioned observations, we model the traffic flow using queuing theory. One way of applying queuing theory can be using exact count of the vehicles in motion. If we have to count the vehicles, it is important to track the vehicles accurately. However, it is difficult to accurately track vehicles in complex scenarios~\citep*{SHBae, Choi_2016_CVPR, KCFTracker, zhou2009object, AMilan1, AMilan2, Zhang_2017_CVPR, yang2011recent}. Thus, we represent the foreground moving objects in terms of temporal clusters. We want smaller vehicles to be composed of lesser number of clusters as compared to the  bigger vehicles. This ensures that if the clusters are used as the elements of the queue, time to cross a typical signal can accurately modeled using departure rate. Consider a bus and a car crossing a particular signal as shown in Fig.~\ref{fig:CLUSTER_VS_VEHICLE}. The car takes lesser time to cross, whereas the bus usually takes more time to cross due to its larger size. By representing clusters as elements of the queuing system, $\lambda$ and $\mu$ can be approximated accurately. As the clusters follow the characteristics of a typical queuing system, this also can be used for understanding and managing the traffic.

\subsection{Related work}
Nowadays, real-time traffic information is collected from various sources like traffic cameras, road sensors and crowd sensing data through users~\citep*{MFathy,Zhang_2017_CVPR, ARAGHI20151538}. Authors of ~\citep*{AJanecek, collotta2015novel} discusses the traffic management using cellular technologies. Global Positioning Systems (GPS) based technologies are used in the work of ~\citep*{MRohani, WLi}  for traffic management. Congestion is one of the key issues in most of the traffic management mechanisms. The research work discussed in~\citep*{ZCao, JPark, FTerroso-Saenz, SWang, wen2010intelligent} focus on congestion analysis and control. Some of the recent work~\citep*{dresner2008multiagent, MHausknecht} discuss traffic management using communication between autonomous vehicles for signal free traffic.  Some of the existing research work~\citep*{ZCao, MHausknecht,JZhao} also discuss signal management at junctions. However, existing methods handle the traffic signal management using the information available from other sources using some communication mechanism. To the best of our knowledge, no such work exists that adopts queuing theory to predict signal duration using a computer vision based approach. Therefore, we have proposed a queuing theory guided expert system for scheduling of traffic signals based on unsupervised learning of temporal clusters.

\subsection{Motivation and Contributions}
We have drawn motivation about this research work by observing how a traffic guard/person handles the flow in a typical 4-way junction. It has been observed that, duration of green signal is usually decided based on the queue from traffic inflow, rather than exactly counting the number of vehicles. The vehicles usually move one behind the other in normal circumstances. This has helped us to model traffic flow by a typical queuing system where objects move in first-come-first-served (FCFS) fashion. With DPMM guided temporal clustering of moving pixels, we can develop tracklets corresponding to the moving objects. This information can be used for modeling the traffic flow in an M-way traffic junction. Based on the above mentioned facts, we list-down a few assumptions and motivations that have guided us to build the proposed traffic management framework, (i) tracking in complex environment can be difficult, thus temporal clustering can be used to represent the traffic movement at a coarse level,  (ii) understanding the dynamic nature of traffic in one direction can provide valuable insight into the overall traffic signal management problem, (iii) the above understanding may help to build a model for complex junctions, and (iv) typical queuing model can be applied to understand the traffic behavior at junction. Motivated by the above facts, we have made the following contributions:
\begin{enumerate}[(i)]
\item We have developed a method to obtain short trajectories (tracklets) with the help of machine learning based DPMM guided temporal clustering and used them to learn traffic behavior at junction.
\item A queuing theory based model has been proposed for understanding traffic state at junctions and to dynamically predict the signal duration in a unidirectional traffic flow, thus making the building block of traffic intersection management for expert systems such as ITS.
\end{enumerate}
The rest of the paper is organized as follows. In Section~\ref{sec:Method}, we describe the background on tracklet generation and formulation for the proposed method. In Section~\ref{sec:Experiments}, we discuss the experiments and discussion of the results. In Section \ref{sec:Conclusion}, we conclude the work with our insight into the future directions of the present work.
\section{Method}
\label{sec:Method}
 \subsection{Background}
    In order to develop unsupervised method for managing traffic flows, it is important to learn the vehicles in motion that fills the space on the road. It has been found that DPMM based models are highly popular for unsupervised learning of clusters~\citep*{LDA, REmonet, WHu, Kuettel, XSun, YeeWhyeTeh, wang2011trajectory}. Firstly, we introduce some terminologies used in this paper. We use observation or data to represent pixels. Topic or cluster denotes a distribution of data and it will be associated with a label. Our proposed model is non-parametric in nature. A parametric model has a fixed number of parameters, while in non-parametric models, parameters grow in number with the amount of data. This characteristic is essential as it can learn more clusters when more objects arrive in the area of interest. 

The model can be mathematically expressed as in ~(\ref{equation:DPM1}-\ref{equation:DPM4}). $z_i$ is a discrete random variable taking one of cluster labels $k$ for the observation $x_i$, where $x_i$ is the random variable representing $i^{th}$ observation such that $i = 1 \cdots N$ with $N$ being the number of observations and $k = 1 \cdots K$ with K being the number of clusters. $\pi$ is a vector of length $K$ representing the probability of $z_i$ taking the value $k$ otherwise called mixing proportion. $\theta_k$ is the parameter of the cluster $k$ and $F(\theta_{z_i})$ denotes the distribution defined by $\theta_{z_i}$. $\alpha$ denotes the concentration parameter and its value decides the number of clusters formed. Firstly, we pick $z_i$ from a Discrete distribution given in~(\ref{equation:DPM1}) and then generate data from a distribution parameterized by $\theta_{z_i}$ as given in~(\ref{equation:DPM2}). Parameter $\pi$ is derived from a Dirichlet distribution as given in~(\ref{equation:DPM3}) and $\theta_k$ is derived from distribution $H$ of priors as represented in~(\ref{equation:DPM4}). The model is graphically~\citep*{GraphicModel} presented in Fig.~\ref{fig:BUILDING_BLOCK}(b). 

\begin{equation}
z_i|\pi  \sim  \mbox{Discrete}(\pi)
\label{equation:DPM1}
\end{equation} 

\begin{equation}
x_i|z_i, \theta_k  \sim  F(\theta_{z_i})
\label{equation:DPM2}
\end{equation} 

\begin{equation}
\pi = (\pi_1, \cdots ,\pi_K)|\alpha  \sim  \mbox{Dirichlet}(\alpha / K, \cdots, \alpha / K)
\label{equation:DPM3}
\end{equation} 

\begin{equation}
\theta_k|H  \sim  H
\label{equation:DPM4}
\end{equation} 

We extend the model temporally with the following additional assumptions that the features do not change significantly between $t-1$ and $t$, where $t$ represents the time stamp of the frame, i.e., state information does not change significantly between consecutive frames.

If $i^{th}$ pixel belongs to an object in both $(t-1)^{th}$ and ${t}^{th}$ frames, the probability of an observation $x_i^{t}$ belongs to a cluster $z_i^{t-1}$ is  expected to be higher than it belongs to another cluster. This implies, cluster parameters are approximately equal between successive frames, i.e., $\theta_k^{t}  \approx \theta_k^{t-1} $. However, they may not be exactly same. If Gibbs sampling~\citep*{Neal2000} is performed using  $\theta_k^{t-1}$ as a prior for the ${t}^{th}$ frame, not only the convergence becomes faster, but also the cluster labels can be maintained between consecutive frames. The rationale behind using only one iteration per frame is that, even if all the observations do not get clustered correctly in the current frame, they are essentially done in  subsequent frames. Thus, the temporal clustering model can be expressed using ~(\ref{equation:ITCM1}-\ref{equation:ITCM4}). 

\begin{equation}
z_i^t|\pi^t  \sim  \mbox{Discrete}(\pi^t)
\label{equation:ITCM1}
\end{equation} 

\begin{equation}
x_i^t|z_i, \theta_{z_i^t}  \sim  F(\theta_{z_i^t})
\label{equation:ITCM2}
\end{equation} 

\begin{equation}
\pi^t = (\pi_1^t, \cdots ,\pi_K^t)|\alpha,\pi^{t-1}  \sim  \mbox{Dirichlet}(\alpha / K^t, \cdots, \alpha / K^t)
\label{equation:ITCM3}
\end{equation} 

\begin{equation}
\theta_k^t|H,\theta_k^{t-1}  \sim  H
\label{equation:ITCM4}
\end{equation}    

Here, $x_i^t (i = 1 \cdots N)$ corresponds to the data at time $t$ and $z_i^t(i = 1 \cdots N)$ corresponds to the latent variable representing  cluster labels, taking one of the values from $k = 1 \cdots K^t$. $N$ is the number of data points and $K^t$ is the number of clusters, $\pi^t$ is a vector of length $K^t$, $\pi_k^t$ represents the mixing proportion of data among clusters, $\theta_k^t$ is the parameter of the cluster $k$, and $F(\theta_{z_i^t})$ denotes the distribution defined by $\theta_k^t$. The difference from DPMM expressed using (\ref{equation:DPM1}-\ref{equation:DPM4}) is the conditional dependency of $\pi^t$ and $\theta_k^t$ on $\pi^{t-1}$ and $\theta_k^{t-1}$, respectively.

\subsection{Formulation for Tracklets}
Optical flow is commonly used to track moving objects~\citep*{REmonet, Kuettel}. We assume that optical flows belonging to foreground pixels follow DPMM~\citep*{Rasmussen2000}. Therefore, initially we extract optical flow features to identify the pixels that are in motion. After background subtraction using a Mixture of Gaussian~\citep*{kaewtrakulpong2002improved} model, clustering has been  applied on these selected pixels using the inference scheme~\citep*{Neal2000} as given in~(\ref{eq_DPMM_1}). The scheme finds out the cluster labels ($k = 1 \cdots K$) for each of the pixels by representing an observation ($x_i$) using ($x, y, \rightarrow $) in a typical DPMM model, where $(x,y)$ represents position of optical flow vector and $\rightarrow$ represents the quantized direction. $x_{-i}$ and $z_{-i}$ represent the respective set of random variables excluding $x_i$ and $z_i$, respectively. $\theta_{k}$ represents the parameters (mean and covariance) of the cluster $k$ and $\theta_{k_{-i}}$ represents the set of parameters of $K$ clusters  excluding the $i^{th}$ observation. Euclidean distance (ED) has been used to measure the distance of $x_i$ from the center of cluster $k$.

\begin{equation}
    \label{eq_DPMM_1}
p(z_i = k | z_{-i},x_{-i},\theta_{k_{-i}},\alpha)
= \begin{cases}
b \times \frac{\alpha}{n_{-i} + \alpha},\text{if $k = K + 1$;}\\
b \times e^{-ED} \times \frac{n_{k_{-i}}}{n_{-i} + \alpha},\text{else.}
\end{cases}
\end{equation}

In order to make sure that cluster labels are maintained temporally, each of the observations ($x_i$) is Gibbs sampled over the optical flow features obtained from the next frame to obtain $z_{i}$ for the next frame. Convergence has been applied temporally using the inference equation (\ref{eq_DPMM_2}) to obtain the final tracklets of the moving clusters.

\begin{equation}
    \label{eq_DPMM_2}
p(z_i^{t} = k |  x_{-i}^*, z_{-i}^*, \theta_{k_{-i}}^*,\alpha)
= \begin{cases}
b \times \frac{\alpha}{n_{-i} + \alpha},\text{if $k = K + 1$;}\\
b \times e^{-ED} \times \frac{n_{k_{-i}}}{n_{-i} + \alpha},\text{else.}
\end{cases}
\end{equation}

In this formulation, $z_{-i}^*$ is different from the $z_{-i}$ discussed earlier. It represents the set of all cluster assignments except for $x_i^{t}$ such that it includes only the latest elements between $z_i^{t-1}$ and $z_i^{t}$ for any $i$. $\theta_{k_{-i}}^*$ is the parameter representing the distribution corresponding to cluster $k$ in the time-stamp $t$ from the set of observations corresponding to $z_{-i}^*$, where $n_{k_{-i}}^*$ is the number of observations in $\theta_{k_{-i}}^*$, and $b$ is  normalization constant. 
      
Since labels of the clusters are maintained across the frames, they create tracklets which can be represented by  $<(x_1,y_1,\rightarrow_1,t_1)$, $(x_2, y_2, \rightarrow_2, t_2)$, ..., $(x_l, y_l, \rightarrow_l, t_l)>$ corresponding to each cluster label, where $<x_j,y_j>$ and $<\rightarrow_j>$ corresponds to position and direction of the cluster center at time $t_j$ and $l$ is the length of the tracklet. Since the tracklets contain time-stamp information, the arrival ($\lambda$) and departure ($\mu$) rates of the clusters on a predefined road segment can be measured and used to model the traffic state.
     
In the context of signal management, we build the model for traffic management in a step-by-step fashion. Initially, we model the traffic signal for unidirectional flow, ignoring flows in other direction. However, we emphasize that, our analysis can be easily extended for modeling signals in $M$-way traffic junctions. 
\subsection{Modeling of Unidirectional Flow}
Consider an $M$-way junction. The traffic is managed by allocating $T_m^c$ time duration for the signal corresponding to the incoming traffic for the $c^{th}$ cycle, where $m = 1 \cdots M$ and $c = 1 \cdots \infty$. One traffic cycle is said to be complete when allocation is completed for all $M$ incoming traffic flows. Cycle time $T^c = \Sigma T_m^c$ for the $c^{th}$ cycle. Our goal is to find optimal time allocation for the $m^{th}$ incoming traffic to achieve optimum throughput. In order to solve this problem, we first solve the problem of optimizing the $m^{th}$ signal. 
    
We can consider $T^c = T_m^c + T_{m_r}^c$, where $T_{m_r}^c$ is the time duration where the signal remains off/red. We want to predict $T_m^{c+1}$, i.e., $m^{th}$ signal duration for the next cycle, based on past information. It has been observed that the departure rate slowly increases when the signal goes green. It becomes steady when most of the vehicles including the vehicles accumulated during the early period start moving. Finally, the rate becomes stable when arrival and departure rates become similar. As per the above observations, we can divide $T_m^c$ primarily into two segments, namely existing queue clearance time ($T_{m_q}^c$) and  traffic free-flow time  ($T_{m_f}^c$). Free-flow time can further be divided as steady state duration and stable duration ($T_s$). Let $\mu$ be the service rate during $T_{m_q}^c$ and $\lambda$ be the arrival rate. We can assume that, between consecutive cycles, the arrival rate does not change significantly. Thus, we need to estimate $\mu$ and $\lambda$ for the next cycle. Let $\mu_a$ \& $\lambda_a$ represent the actual rates and $\mu_e$ \& $\lambda_e$ represent estimated rates. $\mu_a$ and $\lambda_a$ can be assumed to be good estimate of the service and arrival rates for the next cycle as given in~(\ref{eq_1}) and (\ref{eq_2}).

\begin{equation}
    \label{eq_1}
    \mu_e^{c+1} = \mu_a^{c}   
\end{equation}

\begin{equation}
    \label{eq_2}
    \lambda_e^{c+1} = \lambda_a^{c}
\end{equation}

The estimated time for the $m^{th}$ signal can be computed using~(\ref{eq_base}), where $\Delta t^{c+1}$ denotes the error in estimation of $T_{m_q}^{c}$.  The equation needs to satisfy the constraint that predicted throughput is not below the current throughput considering $M$-way signals. For simplicity, we can assume $T^{c+1} = T^{c}$ and $T_{m_r}^{c+1}$ is a non-zero quantity, i.e. fixed cycle duration and nonzero red signal duration. If the above criteria is not met, older value of $T_{m}$ needs to be initialized in the current cycle, i.e., $T_{m}^{c+1} = T_{m}^{c}$.

\begin{equation}
    \label{eq_base}
    T_{m}^{c+1} = T_{m_q}^{c+1} + T_{m_f}^{c+1} + \Delta t^{c+1}    
\end{equation}

The size of the queue that builds up during $T_{m_r}^c$ can be the estimated queue length for the $c+1^{st}$ cycle. Hence queue clearance time can be estimated using~(\ref{eq_4}).

\begin{equation}
    \label{eq_4}
    T_{m_q}^{c+1} = \frac{\lambda_e^{c+1} * T_{m_r}^c}{\mu_e^{c+1}}
\end{equation}

The free-flow time is given in (\ref{eq_5}). It is not sufficient to assume only the queue clearance time. We need to accommodate time for the vehicles getting accumulated when the signal opens. This will help to obtain better throughput. This can be a factor ($\gamma > 1$) of queue clearance time as queue size is proportional to arrival rate. When the arrival rate is more, it is intuitive to allocate more time to the free-flow segment.  We add additional constant time ($T_s$) to make sure that there is a stable free-flow time for each signal to get better estimate of the arrival rate.

\begin{figure}[t]
\begin{center}
   \includegraphics[width=1.0\linewidth]{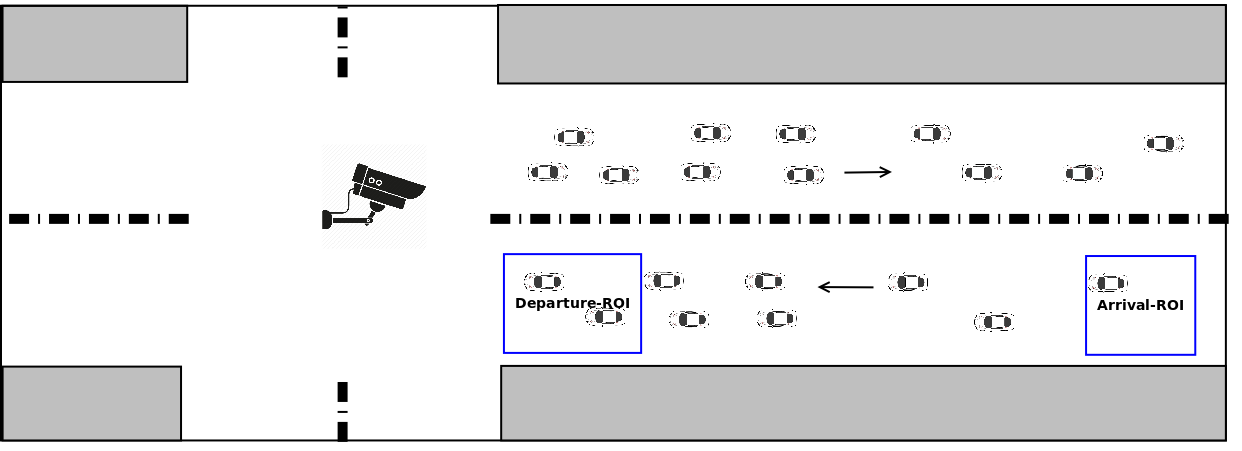}
\end{center}
   \caption{A typical traffic flow representation for a unidirectional flow.}
\label{fig:BOUNDING_BOX}
\end{figure}

\begin{equation}
    \label{eq_5}
    T_{m_f}^{c+1} = T_{m_q}^{c+1}*\gamma  + T_s   
\end{equation}

\begin{equation}
    \label{eq_6}
    \Delta t^{c+1} = \frac{\lambda_a^c - \lambda_e^c}{\lambda_a^c} \times T_{m}^c   
\end{equation}

The challenge is mainly measuring the actual value of $\lambda$ and $\mu$ for the current iteration, $\lambda_a$ and $\mu_a$. If the $\mu$ curve versus queue clearance time is learned, free-flow time period and $\lambda_a$ can be learned. $\mu$ can be learned using non-parametric regression technique based on the data available for a few cycles corresponding to the $m^{th}$ signal. We have used Gaussian Kernel regression for learning $\mu$. If $<t_p$, $\mu_{t_p}>$ represents a data point and P (where $p=1 \cdots P$) such points are known a priori, then, $\mu$ at time $t$ can be found using~(\ref{eq_K_REG_1}), where $t_p$ and $\mu_{t_p}$ represent queue clearance time and service rate. $K(t,t_p) = e^{-\frac{(t_p-t)^2}{2\sigma^2}}$ is the Gaussian kernel~\citep*{takeda2007kernel}.

\begin{equation}
    \label{eq_K_REG_1}
    \mu(t) = \frac{\Sigma _{p=1}^{P}(K(t,t_p)\mu_{t_l})}{\Sigma _{p=1}^{P}(K(t,t_p)}    
\end{equation}

In a unidirectional traffic flow, $\lambda$ can be measured by counting the number of cluster centers entering the bounding box per unit time and $\mu$ can be measured by the number of clusters exiting the bounding box per unit time as depicted in Fig.~\ref{fig:BOUNDING_BOX}. We call these bounding boxes as Regions of Interest (ROI). We call the respective bounding boxes as Arrival-ROI and Departure-ROI, subsequently.

Algorithm \ref{algorithm:1} has been used to learn $\mu$ using the Gaussian kernel regression discussed earlier. It calculates $\mu$ for varying queue clearance time by collecting data points during a few  cycles ($C$). It can be noted that, we may get $P$ data points with lesser number of cycles, i.e., $C << P$. The time required for the $l^{th}$ element to cross the Departure-ROI can be considered as the queue clearance time for a queue length of $l$ since the queue clearance time is only influenced by the number of elements in front of the $l^{th}$ element. This way, there is no need to run the algorithm for $P$ number of cycles to get as many data points. The queue clearance time can be calculated once the time-stamps of the tracklets crossing Arrival-ROI and Departure-ROIs are known. The signal open time-stamp ($t_s$) is known a priori. Once enough number of data points ($P$) are obtained for different queue lengths,  $\mu$-curve can be generated using the Gaussian regression. The result is returned in the form of a list.

\begin{algorithm}
\caption{\textbf{$\mu$ Learner}}
\textbf{Input:} Input video,  $\alpha$ (Concentration parameter for DPMM), $C$ (The number of cycles to learn $\mu$), $t_{max}$ (The upper limit of queue clearance time), $T_m$ (Fixed signal duration for each of the traffic flows) \\
\textbf{Output:} The list $\mu[t_{max}]$, where $\mu[t]$ gives $\mu$ values for different queue lengths $t = 1 \cdots t_{max}$.\\
\textbf{Procedure:}
\begin{algorithmic}[1]
\State Flag each tracklet (calculated as per (\ref{eq_DPMM_1})) with arrival and departure flag along timestamps $t_a$ and $t_d$ on  entering Arrival-ROI and Departure-ROI, respectively;
\State Run the $m^{th}$ signal for duration ($T_m$) by $C$ number of cycles.
\For { each $c$}
\State $tl_{ad}$ = First tracklet with arrival and departure flags;
\State $tl_{ad_{-}}$ = Tracklet before $tl_{ad}$;
\EndFor
\State Create $P$ data points ($t_l,\mu_l$) after calculating $\mu_l = \frac{l}{(t_l - t_s)}$ corresponding to tracklets upto $tl_{ad_{-}}$ for each cycle, where $l$ represents the $l^{th}$ cluster crossing Departure-ROI at time $t_l$ from the  signal start time ($t_s$) in any of the $C$ cycles;
\For {$t_l = 1 \cdots t_{max}$}
\State  Calculate $\mu[t_l]$ as per (\ref{eq_K_REG_1});
\EndFor
\State Return the list $\mu[t_{max}]$;
\end{algorithmic}
\label{algorithm:1}
\end{algorithm} 

Algorithm~\ref{algorithm:2} gives the overall signal prediction mechanism. In the initial part, it learns the $\mu$-curve using Algorithm ~\ref{algorithm:1}. Once $\mu$ is learned, in subsequent cycles, it predicts the signal open duration using~(\ref{eq_base}). It runs the signal with  predicted duration, if it meets the criteria for unidirectional flow, i.e. fixed cycle duration and nonzero red signal duration. If the criteria is not met, the algorithm runs with the previous cycle's signal duration. We have represented it as a function $criteria()$ for future extensibility as the criteria can be for achieving optimal throughput.

In the algorithms, four parameters, namely ($\alpha$, $T_m$, $t_{max}$, $C$) have been used. $\alpha$ only affects the number of clusters/vehicle, not the signal duration estimation. Fixed signal duration $T_m$s can be used for learning the $\mu$ values. $t_{max}$ represents the maximum queue clearance time possible. $C$ is number of cycles to learn departure rate ($\mu$). 

\begin{algorithm}
\caption{Adaptive Signal Duration Predictor}
\textbf{Input:} Input video; $\alpha$ (The concentration parameter for the DPMM), $C$ (The number of cycles to learn $\mu$), $t_{max}$ (The upper limit of queue clearance time), $T_m$ (Fixed signal duration for each of the traffic flows)\\
\textbf{Output:} Predicted Signal duration for $m^{th}$ signal.\\
\textbf{Procedure:}
\begin{algorithmic}[1]
\State Run the $\mu$ Learner as per (\ref{algorithm:1});
\State Measure queue clearance time ($T_{m_q}^{c}$ = $t_d$ of $tl_{ad_{-}}$ - $t_s$); 
\State Calculate departure rate ($\mu_{a}^{c} = \mu[T_{m_q}^{c}]$);
\State Calculate arrival rate ($\lambda_{a}^{c}$ = (\# tracklets crossed Departure-ROI during $T_s$) / $T_s$);
\For {$c = C \cdots \infty$}
\State Set $\mu_e^{c+1}$ and $\lambda_e^{c+1}$ as per (\ref{eq_1}) and (\ref{eq_2});
\State Calculate $T_{m}^{c+1}$ as per (\ref{eq_base});
\If {($criteria() == TRUE$)}
\State Run the signal for $T_{m}^{c+1}$;
\Else
\State Run the signal for $T_{m}^{c}$;
\EndIf
\State Measure queue clearance time ($T_{m_q}^{c+1}$);
\State Calculate $\mu_a^{c+1}$ for $T_{m_q}^{c+1}$;
\State Calculate $\lambda_a^{c+1}$;
\EndFor
\end{algorithmic}
\label{algorithm:2}
\end{algorithm} 

\section{Experiments}
\label{sec:Experiments}
Experiments have been conducted to validate our assumptions and to establish the claim that the traffic flow can be modeled using a queuing theory based approach. We have used two publicly available surveillance video datasets QMUL~\citep*{russell2008multi_QMUL_dataset} and MIT~\citep*{XWang_MIT_Traffic_dataset}. Signal duration prediction experiments have been conducted only using QMUL dataset since the other dataset does not provide visual clue about the actual signal on/off durations.  

\begin{figure*}[t!]
\begin{center}
 \subfigure[QMUL inter-arrival time distribution]{\includegraphics[scale=0.29]{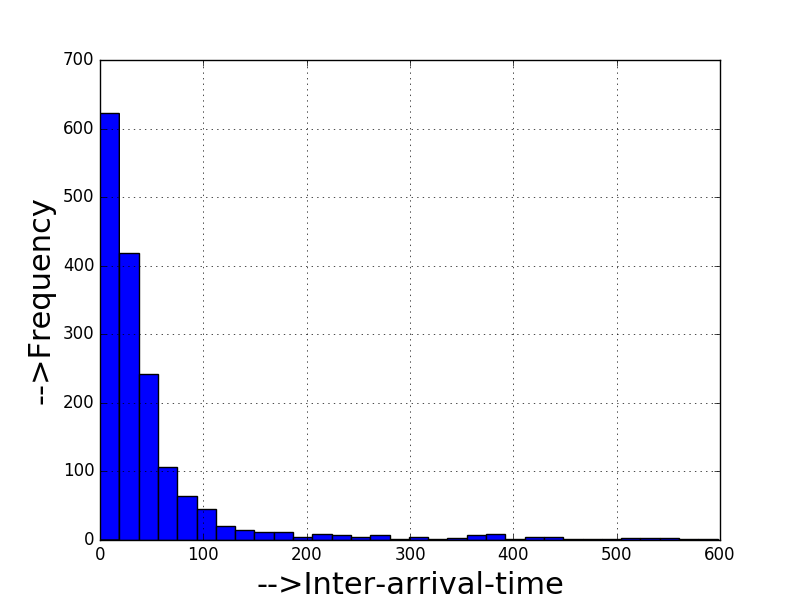}}   
 \subfigure[QMUL inter-departure time distribution]{\includegraphics[scale=0.29]{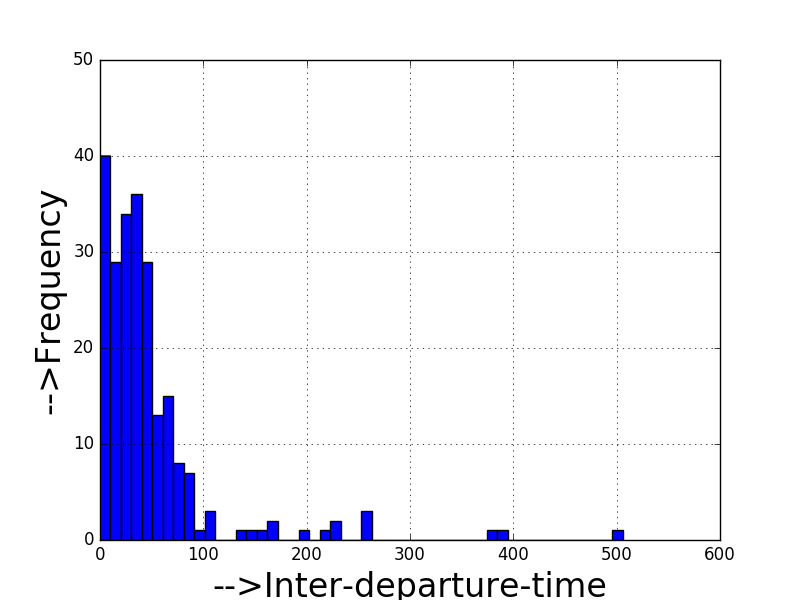}}   
 \subfigure[MIT inter-arrival time distribution]{\includegraphics[scale=0.29]{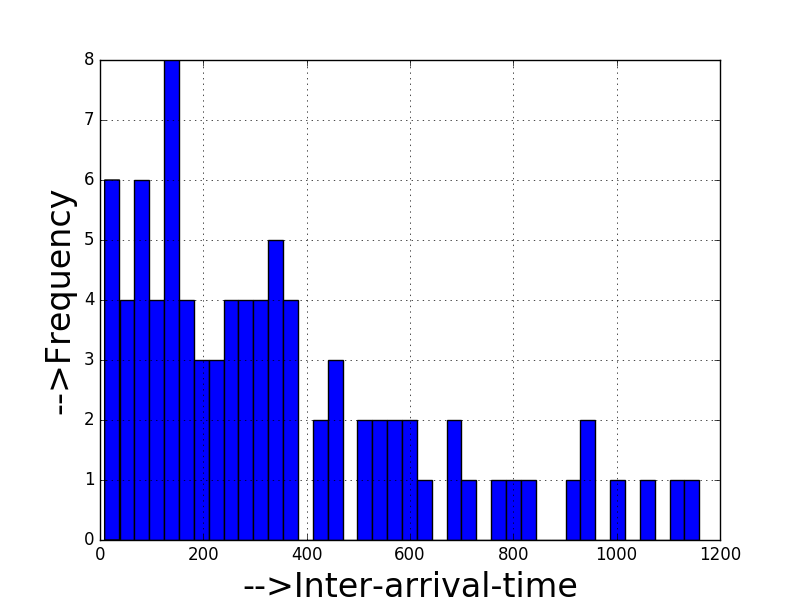}}   
 \subfigure[MIT inter-departure time distribution]{\includegraphics[scale=0.29]{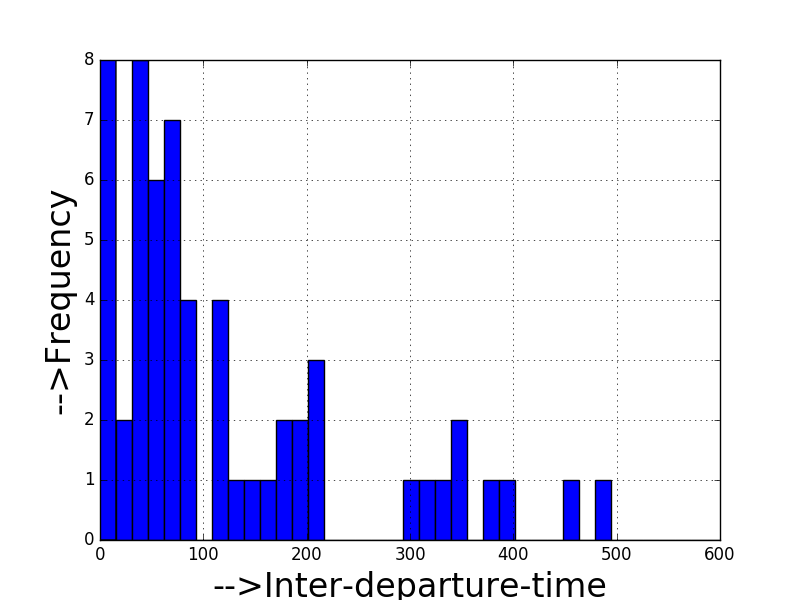}}  
\end{center}
   \caption{Inter-arrival time distributions for two datasets.}
\label{fig:INTER_ARRIVAL}
\end{figure*}

\subsection{Traffic State Analysis}
\begin{figure}[t!]
\begin{center}
 \subfigure[$\lambda$ and $\mu$ plots for a few cycles]{\includegraphics[scale=0.29]{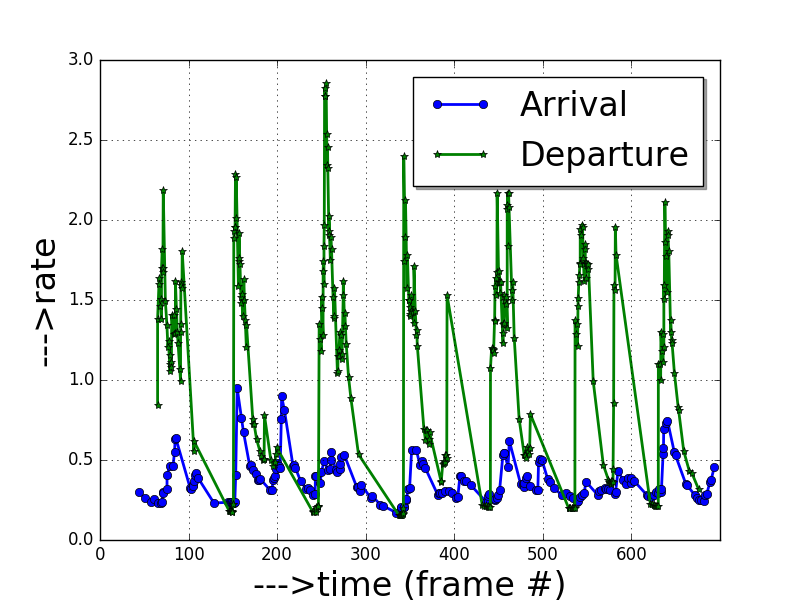}}   
 \subfigure[$\lambda$ and $\mu$ plots for one signal duration]{\includegraphics[scale=0.29]{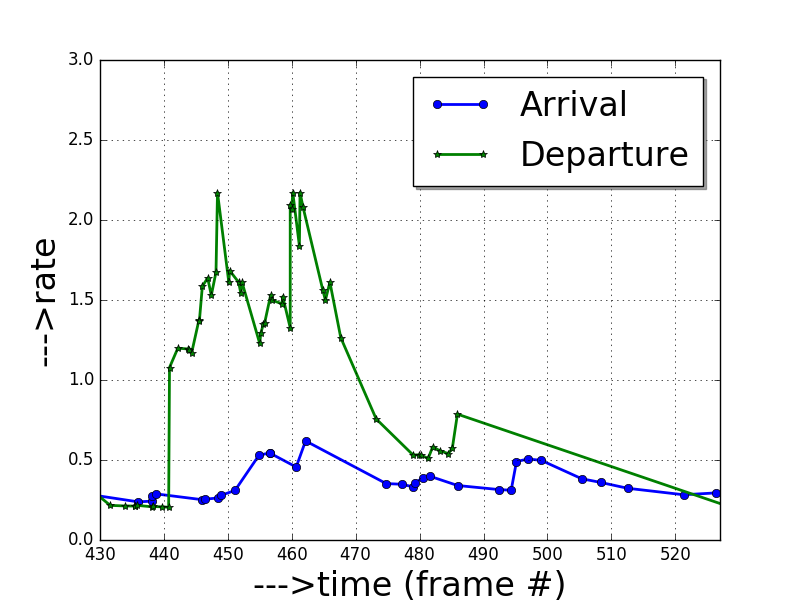}} 
 \subfigure[Clearance time vs. Cumulative departure rate]{\includegraphics[scale=0.19]{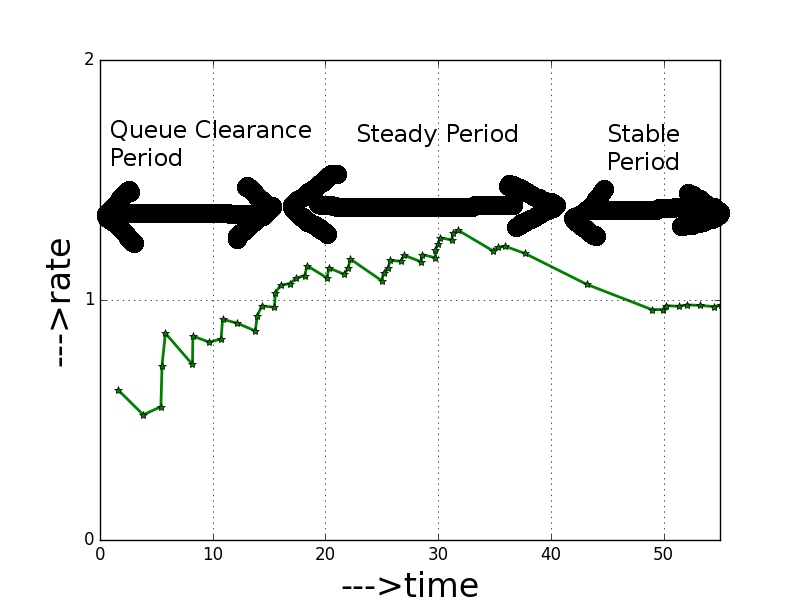}}   
 \subfigure[Cumulative \# Elements vs. Cumulative departure rate]{\includegraphics[scale=0.19]{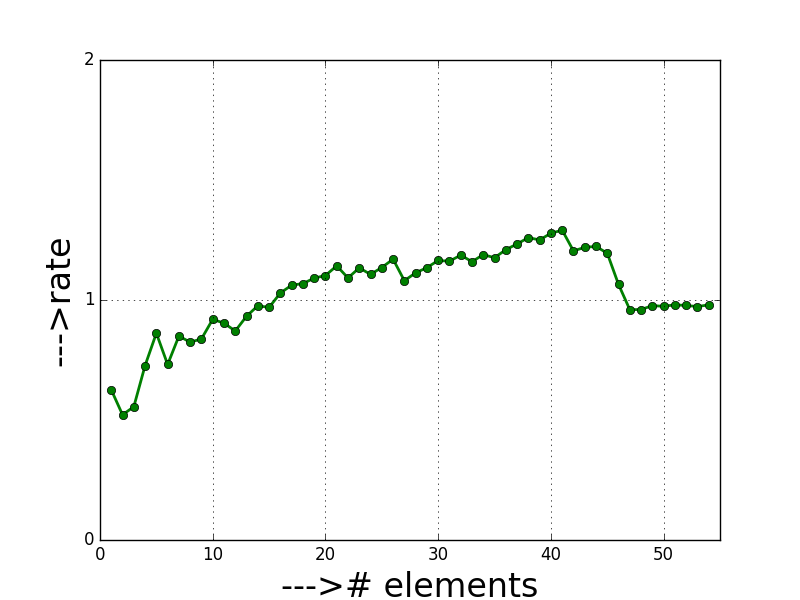}}   
 \subfigure[Cumulative \# of Elements  vs. Clearance time]{\includegraphics[scale=0.19]{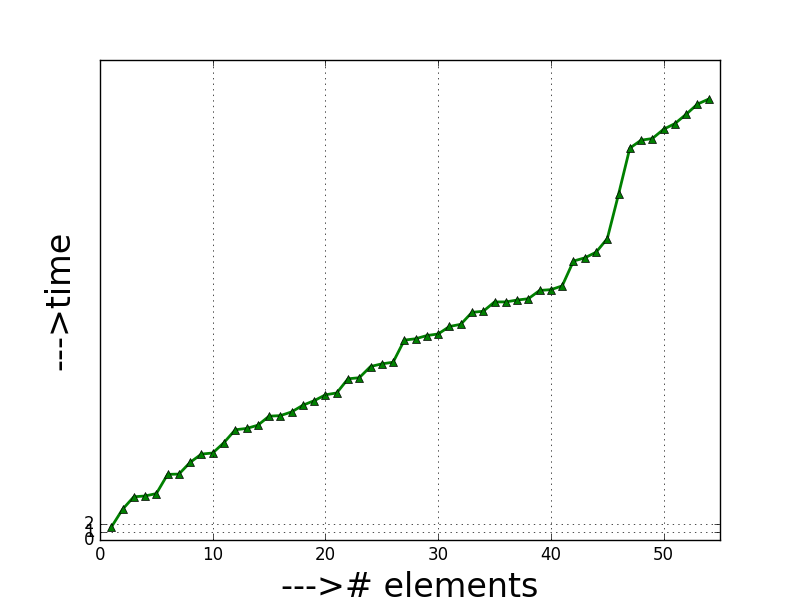}}     
\end{center}  
  \caption{Plot of queuing parameters obtained using the QMUL junction video. (a) Arrival and departure rates for a few cycles. (b) Arrival and departure rates for one of the typical cycles. The spikes in the departure rate correspond to the signal opening time. The arrival rate indicated by blue plots has been found to be steady for majority of the duration. Spikes in certain regions of the arrival curve happen due to the presence of a signal opening before Arrival-ROI which causes an increase in the traffic flow. The relation between cumulative departure rate, cumulative \# of clusters ($n$) that crossed Departure-ROI, and their clearance time are represented in (c), (d), and (e). As discussed in Section 2.3, initial departure rate is slow which is expected. As marked in (c), second segment corresponds to the steady flow period when the vehicles are moving including the vehicles that are accumulated during queue clearance. The final segment corresponds to a stable duration when arrival and departure rates are similar (as if there is not traffic signal). Plot shown in (e) clearly indicates a linear relation between the number of elements in the queue and the queue clearance time.}
\label{fig:ARRIVAL_DEP_RATE}
\end{figure}

As discussed earlier, a queuing model is characterized by the distributions of inter-arrival time, service time and number of servers. Inter-arrival time plotted for a few traffic datasets clearly indicate that it follows exponential distribution as can be verified from Fig.~\ref{fig:INTER_ARRIVAL}. If the time between consecutive occurrences of an event follows exponential distribution with parameter $\delta$, we can express the probability density function  $f(t) = \delta e^{-\delta t}$ for $t >= 0$, otherwise $f(t)=0$. Then, the number of occurrences ($X(t)$) within the interval $t$ has a Poisson distribution with parameter $\delta t$. The mean of the distribution is $E [X(t)] = \delta t $. The expected number of events/unit time is $\delta$ and it is the mean rate at which the events occur. In the case of arrival event, $\delta = \lambda$. Arrivals are said to occur according to a Poisson input process with the parameter $\lambda$. Similarly, the service time also follows an exponential distribution as it can be seen from Fig.~\ref{fig:INTER_ARRIVAL}(b) and Fig.~\ref{fig:INTER_ARRIVAL}(d), where $\delta = \mu$. We consider the number of servers to be one corresponding to the inflow traffic. Fig.~\ref{fig:ARRIVAL_DEP_RATE} shows the arrival and departure rates obtained using the QMUL junction video for a few cycles and their detailed analysis.

\begin{figure*}[t!]
\begin{center}
 \subfigure[Scene snapshot]{\includegraphics[scale=0.31]{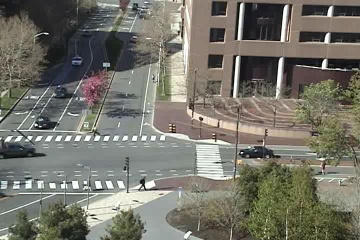}}   
 \subfigure[Clusters snapshot]{\includegraphics[scale=0.31]{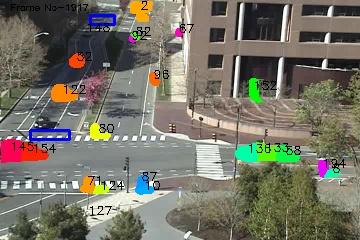}}  
 \subfigure[Tracklet snapshot]{\includegraphics[scale=0.31]{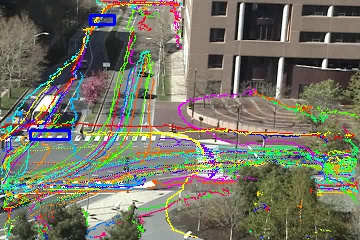}} 
 \subfigure[Freeflow tracklets]{\includegraphics[scale=0.31]{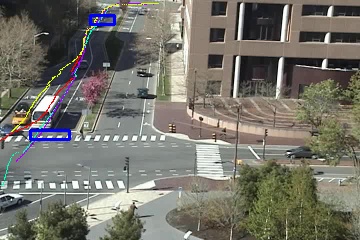}}      
 \subfigure[Queuing tracklets]{\includegraphics[scale=0.31]{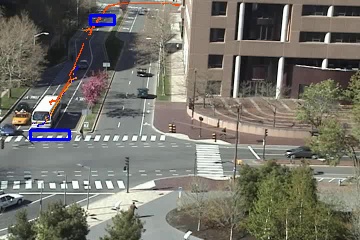}}      
 \subfigure[Queue clearing tracklets]{\includegraphics[scale=0.31]{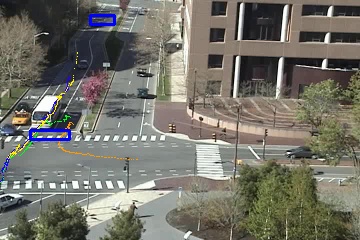}}      
\end{center}
   \caption{Representation of different features obtained using the MIT dataset videos. (a) Represents the snapshots of a scene. (b) Corresponding clusters of the scene. (c) Tracklets found till the current frame. (d-f) Depict different kinds of tracklets in a unidirectional flow.}
\label{fig:TRACKLETS_MIT}
\end{figure*}

It has been found during tracklet analysis that, there are three kinds of tracklets possible in a typical traffic flow. A free-flow tracklet, which is formed when the signal is clear and vehicles need not stop at the signal. Second type is queuing tracklets which are formed when the vehicles are stopped at the signal. Third kind is the queue clearing tracklets. These are formed when the signal turns on and the vehicles start moving. Fig.~\ref{fig:TRACKLETS_MIT} demonstrates all such kinds tracklets with visual marking. 
\subsection{Traffic Signal Management}

\begin{figure}[t!]
\begin{center}
 \subfigure[QMUL scene snapshot]{\includegraphics[scale=0.3]{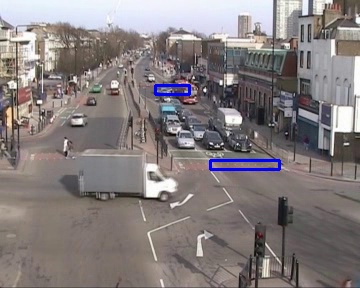}} 
 \subfigure[QMUL $\mu$-curve]{\includegraphics[scale=0.4]{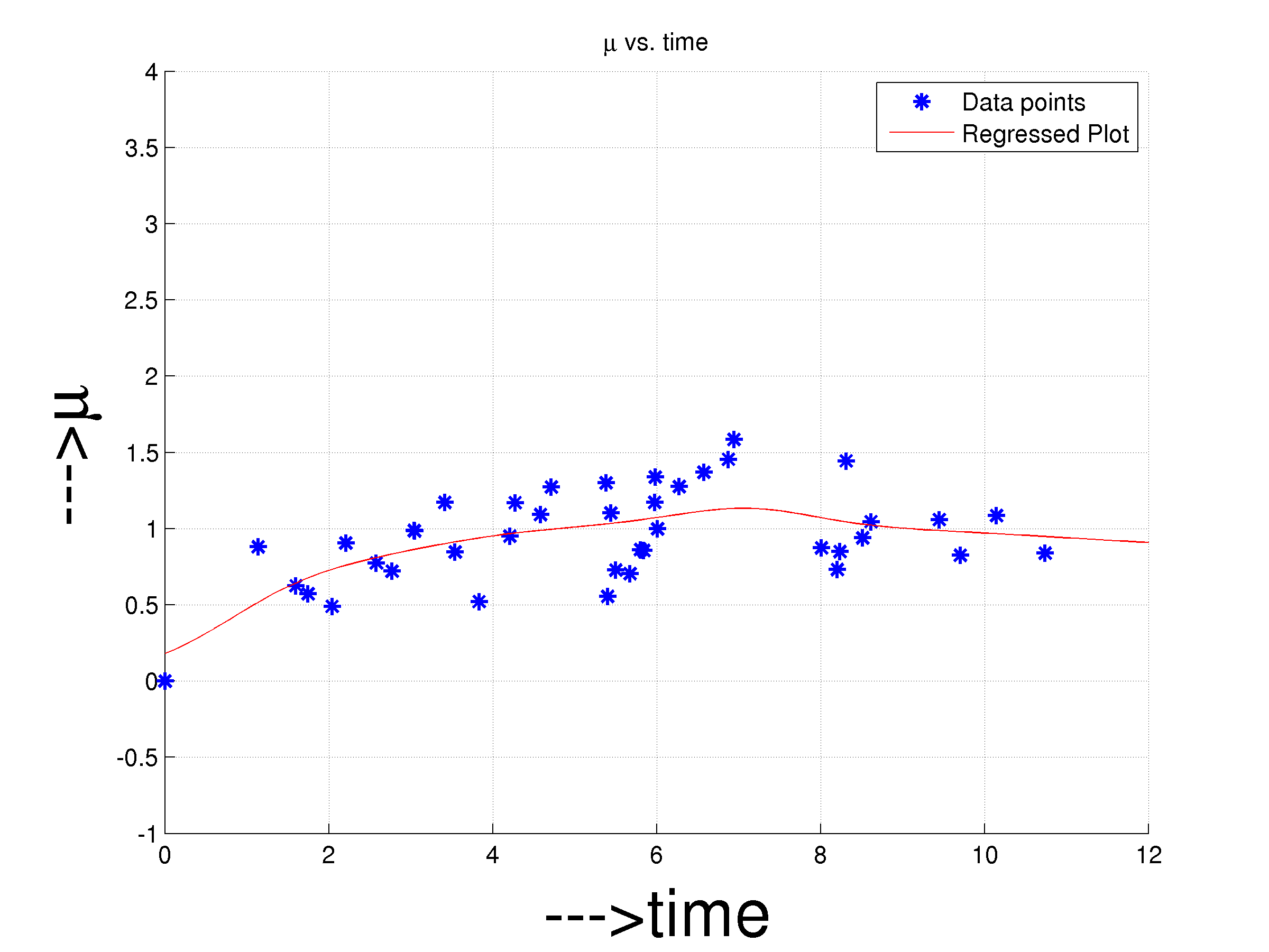}}    
\end{center}
   \caption{QMUL experimental data. (a) The unidirectional flow marked with Arrival-ROI and Departure-ROI. (b)$\mu$-curve learned using the Gaussian regression.}
\label{fig:QML_MU_CURVE}
\end{figure}

\begin{table}
\begin{center}
\caption {Mean Absolute Error (MAE) Table} 
\label{Tab:MAE} 
    \begin{tabular}{| l | l | l | l | l | l |}
    \hline
    c & $T_G$(s) & $T_R$(s) & $T_q^e$ & $T_q^a$(s) & MAE (\%) \\ \hline
    1 & 55 & 38 & - & 05.98 & -\\ \hline
    2 & 56 & 38 & 14.17 & 10.14 & 15.50\\ \hline
    3 & 55 & 43 & 08.91 & 08.31 & 38.50\\ \hline
    4 & 55 & 39 & 09.47 & 10.74 & 02.36\\ \hline    
    5 & 55 & 39 & 13.32 & 15.18 & 01.20\\
    \hline
    \end{tabular}
\end{center}
\end{table} 
Firstly we describe the parameter values ($\alpha$, $T_m$, $t_{max}$, $C$) used in our algorithms. For QMUL dataset shown in Fig.~\ref{fig:QML_MU_CURVE}(a), $\alpha= 0.0000003$ (estimated empirically) produced near-optimal cluster (small object) or set of clusters (large objects) closely representing moving vehicles when the viewing perspective is not changed. We have used a fixed signal duration $T_m=55$s by visually observing the QMUL video. The maximum of queue clearance ($t_{max}$) is found to be 12s for QMUL video for the selected cycles. We have used first four cycles ($C = 4$) to learn the $\mu$. We have used 4 cycles ($C = 4$) to learn $\mu$ using Gaussian regression and the results are shown in Fig.~\ref{fig:QML_MU_CURVE}(b). We have conducted tests to predict the queue clearance time for 5 cycles and the results are presented in Table~\ref{Tab:MAE}, where $c$ denotes the cycle number, $T_G$, $T_R$, $T_q^e$ and $T_q^a$  denote green signal duration, red signal duration, estimated queue clearance time, and ground truth queue clearance time, respectively. It can be observed that the proposed method predicts the queue clearance time with high accuracy. Since the duration of the signal is highly dependent on the queue clearance time, the method can be used for managing traffic signals. In order to illustrate the point, we have considered one of the cycles and observed the predicted cycle duration. We consider $c = 4$, thus we want to predict the time duration for the $5^{th}$ cycle. As per the ground truth measurement, $T_{m_q}^4 = 10.74$, $\lambda_a^4 = 0.30$, $\lambda_a^4 = 0.25$, $T_{m_r}^4 = 42$ and we assume $\gamma = 2$. Thus,

\begin{align*}
T_m^{c+1}
&= T_{m_q}^{c+1} + T_{m_f}^{c+1} + \Delta t^{c+1} \\
&= T_{m_q}^{c+1} + (T_{m_q}^{c+1}*\gamma + T_s) + \Delta t^{c+1} \\
&= \frac{\lambda_a^{c} * T_{m_r}^c}{\mu_e^{c+1}} + T_{m_q}^{c+1}*\gamma + T_s + (\frac{\lambda_a^c - \lambda_e^c}{\lambda_a^c}) * T_{m_q}^c \\
&= \frac{0.3 * 42}{0.9463} + T_{m_q}^{c+1}*2 + 20 + \frac{(0.3 - 0.25)}{0.3} * 10.74\\
&= 13.32 + (13.32*2 + 20) + (\frac{0.3 - 0.25}{0.3}) * 10.74\\
&= 13.32 + (13.32*2 + 20) + 1.79\\
&= 61.75\\
\end{align*}

It may be noted that, if the arrival rate increases, the signal duration needs to be proportionately increased depending on the value of $\gamma$. In the above example, the arrival rate has increased from ($0.25$ to $0.3$). Hence the $\Delta t^{c+1}$ term is positive and that corrects the error from the last cycle. The predicted duration is $61.75$s as compared to the ground truth value of $55$s, i.e., our algorithm accommodates the increase in the arrival rate in predicting next signal duration.  Similarly, when the arrival rate becomes less as compared to the previous cycle, $\Delta t^{c+1}$ becomes negative and the duration is proportionately reduced.  This way in each cycle,  estimation error is corrected with the $\Delta t^{c+1}$ term. The results show that the method can be used for signal duration prediction. Though, this experiment is applied only on a unidirectional traffic flow, the same formula can be applied for every other flows. An objective function can be developed for calculating the throughput. Once a cycle in an M-way junction is over, if the predicted throughput becomes less than the current throughput, then the signal duration for the current cycle is to be repeated for the next cycle. This way, the signal duration is constrained for each of the M-traffic flows as increasing the signal duration for a particular flow beyond a certain threshold may reduce the overall throughput. In a steady traffic condition, the signal duration will be stabilized proportionately to the arrival rate.

\subsection{Comparative Analysis}
\begin{figure}[t!]
\begin{center}
 \subfigure[Measurement vs. Ground Truth]{\includegraphics[scale=0.29]{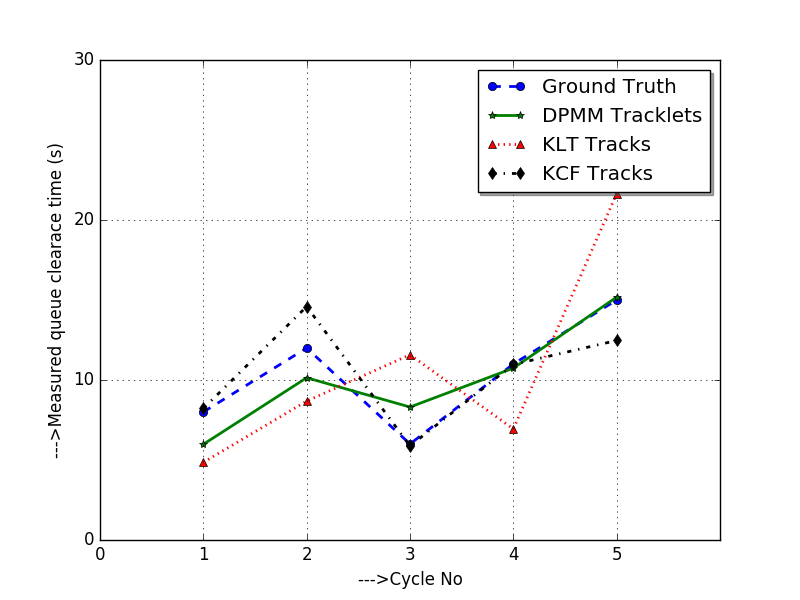}}    
 \subfigure[Prediction vs. Ground Truth]{\includegraphics[scale=0.29]{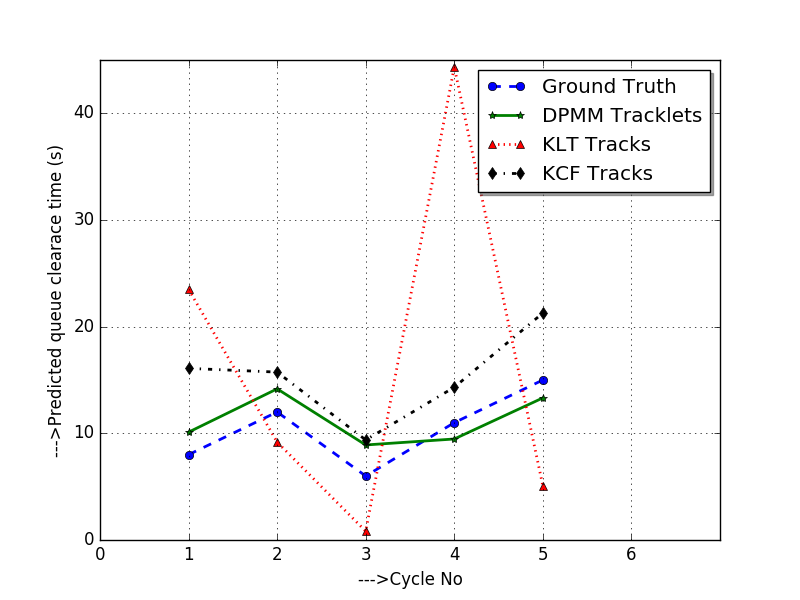}}   
\end{center}
   \caption{Comparison of the proposed method of DPMM tracklets with KCF tracks and KLT tracks. (a) Comparison plots of the measurement accuracies against the ground truths for five cycles. (b) Comparison plots of the prediction accuracies for five cycles.}
\label{fig:COMPARISON}
\end{figure}

We have compared the effectiveness of the proposed DPMM guided feature tracker in predicting signal duration with two existing trackers, namely Kernel Correlation Filters (KCF)~\citep*{KCFTracker} tracker and Kanade-Lucas-Tomasi (KLT) based feature tracker~\citep*{tomasi1991detection}. The comparison results are shown in Fig.~\ref{fig:COMPARISON}. As we have found that KCF fails to track incoming traffic objects accurately while the objects approach the Arrival-ROI, we have reinitialized the tracks for experimental evaluation. This helps to examine whether object tracks from the best algorithm can be used as elements of the proposed queuing model. It has been observed that accurate trackers can provide better measurement accuracy. However, they often fail to predict the queue clearance time accurately as compared to DPMM tracklet features. This is because, spacial occupancy by the vehicles on the road is not taken into consideration when the objects are used as the elements during the learning of the queuing parameters.

 Our initial assumption was that KLT features could be used since bigger vehicles generate more number of features. In case of KLT, the feature tracks are better as compared to KCF. However, KLT tracks have produced lesser measurement accuracy and prediction accuracy as compared to our proposed feature. The reason for low accuracy of prediction and measurement has been found to be the non-correlation between size of the vehicles and number of feature points. The number of feature points are varied even for similarly sized vehicles. As per our observation, the number of feature points depends on the appearance of vehicle than the size. Though DPMM-based tracks are also noisy, we have found that the number of clusters are similar for equal sized vehicles.  Hence DPMM guided tracklets perform better in terms of measurement and prediction accuracies as compared to other trackers. Our feature considers the space occupied on the road as the clusters are of similar size. The clusters (elements) flow on the road (queue) one behind the other, thus giving better results during the prediction.  

\subsection{Discussions and Limitations}
In our experiments we have shown the signal prediction only for 4 cycles due to non availability of cycles of fixed length. We found five consecutive cycles with fixed length (approximately 94s) and hence it is shown in the results. However, the results give an insight into its applications for unmanned intersection management to achieve optimal throughput. 

There are a few limitations of the present method. In order to use the proposed method, it is important to keep the camera at an elevated position to get a top view or near top view of the scene. This will make sure that the visibility of the queue is maximized. Currently, the datasets that have been used in our experiments do not fully support the above requirement. In addition to that, the proposed method depends on the optical flow features. Therefore, robust optical flow estimation is a prerequisite for the success of the proposed model. 
\section{Concluding Remarks and Future Directions}
\label{sec:Conclusion}
With temporal clustering, tracklets are created corresponding to moving objects and they become the elements of the queuing system. Tracklets are used for finding out the arrival and the departure events of vehicles in the queue. A queuing model is applied to learn arrival ($\lambda$) and departure ($\mu$) rates of the vehicles. Learned information is used for predicting signal duration for the next cycle. This method provides an unsupervised way of predicting the signal duration to maximize the throughput. The method has been verified using standard video dataset and comparison reveals that it can be used for predicting the signal duration in traffic junctions. As a future work, we would like to make datasets for junctions that can be used for traffic analysis and signal management. Also, we aim to extend our work considering all possible flows in M-way traffic junctions. 

\section*{Acknowledgment}
\textbf{Funding:} This study is not funded from anywhere.\\
\textbf{Conflict of interest:} The authors declare that there is
no conflict of interest regarding the publication of this paper.\\
\textbf{Ethical approval:} This article does not contain any
studies with human participants or animals performed by any of the
authors. \\
\textbf{Informed consent:} Informed consent was obtained from all
individual participants included in the study.
\section*{References}

\bibliographystyle{apalike}

\end{document}